%% file: main.tex
\documentclass[10pt,twocolumn,letterpaper]{article}

\usepackage[pagenumbers]{cvpr} 

\input{preamble}
\definecolor{cvprblue}{rgb}{0.21,0.49,0.74}
\usepackage[pagebackref,breaklinks,colorlinks,allcolors=cvprblue]{hyperref}
\usepackage{pifont}
\usepackage{wrapfig}
\usepackage{xcolor}
\usepackage{bm}
\usepackage{multirow}

\def\paperID{1104} 
\def\confName{CVPR}
\def\confYear{2026}

\title{Learning from Dense Events: Towards Fast Spiking Neural Networks Training \\
via Event Dataset Distillation}
\author{Shuhan Ye \and Yi Yu \and Qixin Zhang \and Chenqi Kong \and Qiangqiang Wu \and Kun Wang \and Xudong Jiang\\
Nanyang Technological University\\
{\tt\small SHUHAN006@e.ntu.edu.sg}
}

\begin{document}
\maketitle

\input{sec/0_abstract}    
\input{sec/1_intro}
\input{sec/2_formatting}

\input{sec/3_finalcopy}
{
    \small
    \bibliographystyle{ieeenat_fullname}
    \bibliography{main}
}


\end{document}

%% file: sec/0_abstract.tex
\begin{abstract}
Event cameras sense brightness changes and output binary asynchronous event streams, attracting increasing attention. 
Their bio-inspired dynamics align well with spiking neural networks (SNNs), offering a promising energy-efficient alternative to conventional vision systems.
However, SNNs remain costly to train due to temporal coding, which limits their practical deployment.
To alleviate the high training cost of SNNs, we introduce \textbf{PACE} (Phase-Aligned Condensation for Events), the first dataset distillation framework to SNNs and event-based vision.
PACE distills a large training dataset into a compact synthetic one that enables fast SNN training, which is achieved by two core modules: \textbf{ST-DSM} and \textbf{PEQ-N}.
ST-DSM uses residual membrane potentials to densify spike-based features (SDR) and to perform fine-grained spatiotemporal matching of amplitude and phase (ST-SM), while PEQ-N provides a plug-and-play straight through probabilistic integer quantizer compatible with standard event-frame pipelines. 
Across DVS-Gesture, CIFAR10-DVS, and N-MNIST datasets, PACE outperforms existing coreset selection and dataset distillation baselines, with particularly strong gains on dynamic event streams and at low or moderate IPC.
Specifically, on N-MNIST, it achieves \(84.4\%\) accuracy, about \(85\%\) of the full training set performance, while reducing training time by more than \(50\times\) and storage cost by \(6000\times\), yielding compact surrogates that enable minute-scale SNN training and efficient edge deployment.

\end{abstract}

%% file: sec/1_intro.tex
\section{Introduction}
Event cameras such as the Dynamic Vision Sensor (DVS) sense brightness changes and produce binary asynchronous event streams with microsecond latency, extreme dynamic range, and inherently sparse spatiotemporal structure \citep{lichtsteiner2008dvs,Gallego2020Survey}. 
In parallel, spiking neural networks (SNNs), the biologically inspired third generation of artificial neural networks \citep{m:97}, offer event-driven perception, and energy efficiency \citep{hu2023fast, subbulakshmi2021biomimetic}. 
In SNNs, spikes encode when information arrives, computation proceeds asynchronously, and activity remains sparse by design \citep{LIF, zhou2023computational, hu2021spiking}.
The combination of DVS and SNNs has therefore emerged as a compelling blueprint for neuromorphic vision pipelines that complement frame-based systems and can even replace them in latency or power critical scenarios \citep{Rueckauer2017,Wu2018STBP,ckd, qian2025ucf}.

Despite this promise, training SNNs on long, temporally encoded event streams is still computationally expensive, because temporal coding requires forward accumulation over \(T\) time steps and spatiotemporal backpropagation through time \citep{Wu2018STBP}, so the training complexity grows linearly with the number of time steps, \textit{i.e.,} \(\mathcal{O}(T)\) \citep{guo2023efficient, zhou2024direct, feng2025efficient}. 
Meanwhile, many DVS ``dynamic image'' datasets are generated by inducing saccade-like camera motions while replaying static images on a monitor \citep{Orchard2015Saccade} , which produces sharp contours but only limited temporal variation and substantial redundancy. 
In practice, SNNs further discretize these streams into a small number of time bins, increasing redundancy while stabilizing training. Together, these factors make learning on DVS streams both computationally expensive and data inefficient, which limits the scalability and practical deployment of neuromorphic systems but also suggests that carefully distilled synthetic event sets could replace the full streams for training.

Dataset distillation (DD), also known as the dataset condensation, offers a natural way forward \citep{sun2024diversity, liu2024dataset}. The goal is to synthesize compact, information-dense surrogates that allow models to approach the performance of full datasets while sharply reducing storage and compute. Early work formulates DD as \emph{gradient matching} between models trained on real and synthetic data \citep{Zhao2021GM}. Later methods focus on matching \emph{training trajectories} in order to better capture optimization dynamics \citep{Cazenavette2022Traj}. More recent approaches perform \emph{distribution matching} or learn kernel-based surrogates and obtain partial guarantees in simplified settings \citep{Zhao2023IDM,Nguyen2021KIP}. However, DD for \emph{event data} remains unexplored. Neuromorphic signals are sparse, polarity-signed, and strictly time-causal, and SNN learning relies on non-differentiable spikes with cross-scale temporal dependencies, which together make direct transfers of image-domain DD methods unreliable \citep{Gallego2020Survey}.


In this paper, we present \textbf{PACE} (Phase-Aligned Condensation for Events), the first event-based dataset distillation framework designed for fast SNN training on event streams. 
PACE distills a large training set into a compact synthetic one that still drives high-accuracy SNNs. This is realized by two core modules, \textbf{ST-DSM} and \textbf{PEQ-N}. 
ST-DSM uses residual membrane potentials to densify spike-based features (SDR) and performs fine-grained spatiotemporal matching of amplitude and phase (ST-SM) in both feature and time domains. PEQ-N provides a plug-and-play straight-through probabilistic integer quantizer that outputs hard integer frames in the forward pass while keeping gradients and compatibility with standard event-frame pipelines in the backward pass. 
Across DVS-Gesture, CIFAR10-DVS, and N-MNIST, PACE consistently outperforms existing coreset selection and dataset distillation baselines, with the largest gains on dynamic event streams and at low or moderate IPC. On N-MNIST with IPC\(=1\), it reaches \(84.4\%\) accuracy, which is about \(85\%\) of the full training set performance, while reducing training time by more than \(50\times\) and storage cost by \(6000\times\). These results show that PACE yields compact, accurate surrogates that enable minute-scale SNN training on neuromorphic streams and are suitable for efficient edge deployment.


Our contributions are summarized as follows:
\begin{itemize}
    \item We present the first dataset distillation framework for SNNs on event streams and establish a standardized benchmark, introducing the event-native method \textbf{PACE}.
    \item We make spatiotemporal feature matching tractable by densifying spikes with residual membrane potential, and we align real and synthetic streams via CF in feature space and FFT along time within \textbf{ST-DSM}.
    \item We maintain pipeline compatibility while preserving gradients by introducing \textbf{PEQ-N}, a straight-through probabilistic event quantizer that outputs hard integer frames, and a time-expanded condensation objective on frozen teacher features that updates only the synthetic data.
    \item We demonstrate large reductions in storage and training time while maintaining high accuracy across SNN backbones and datasets, enabling efficient SNN training under tight memory and latency budgets.
\end{itemize}

\section{Related Work}
\subsection{Dataset Distillation}
{Dataset distillation}
Dataset distillation (DD) compresses a large training set into a compact synthetic one while preserving downstream accuracy~\citep{wang2018dataset}, typically by optimizing a small set of synthetic examples so that training on them approximates using the full data.
Early methods align point-wise features with Euclidean losses, minimizing $\lVert f(x)-f(\tilde{x})\rVert_2^2$ between real and synthetic samples~\citep{wang2022cafe,zhou2022nfr}. A second line matches distributional statistics with Maximum Mean Discrepancy (MMD), which has been used for dataset condensation~\citep{zhang2024m3d,zhao2022dm,Zhao2023IDM}. In parallel, gradient-matching approaches align gradients induced by real and synthetic batches on the same network~\citep{Zhao2021GM}, but they are slow and memory-hungry, and backpropagation through time makes this even more severe for SNNs. Distribution-matching variants replace gradient alignment with feature-distribution alignment to improve efficiency and stability~\citep{zhao2022dm}. However, most feature-matching and MMD objectives mainly constrain the \emph{magnitude} of statistics and largely ignore \emph{phase} across channels and time. Neural Characteristic Function Matching (NCFM) instead casts distribution matching as a min–max game and uses a neural characteristic function to match both amplitude and phase, providing a principled alternative to MSE- and MMD-based surrogates~\citep{wang2025ncfm}.


\subsection{Spiking Neural Networks (SNNs)}
SNNs compute with discrete spikes and leverage temporal coding, yielding energy and latency gains on neuromorphic hardware~\citep{merolla2014truenorth}. High accuracy is achieved either by \emph{ANN$\rightarrow$SNN conversion}~\citep{rueckauer2017conversion,deng2021optimal,bu2022optimalconv}, which calibrates activations and simulation length to cut conversion error and time steps, or by \emph{direct training} with surrogate gradients, where STBP~\citep{Wu2018STBP} enables deep training and later refinements tune surrogate shapes~\citep{li2021dspike,wang2023asgl}, add time-aware normalization (tdBN~\citep{zheng2021tdbn}), and encourage few-step inference (TET~\citep{deng2022tet}). SNN inputs are either static frames unfolded over $T$ steps or event streams binned into temporal grids~\citep{lichtsteiner2008dvs,amir2017dvsgesture}; this unrolling makes training cost scale with $T$ under BPTT and surrogate gradients, motivating \emph{dataset distillation} to reduce expensive updates while preserving accuracy. Crucially, event data are sparse and polarity-asymmetric and use integer or binary grids, and SNNs carry internal state (\textit{e.g.,} membrane potentials and thresholds), so RGB-oriented distillation objectives built around continuous intensities and CNN features may transfer poorly. While dataset distillation is well studied on frame-based benchmarks, there is, to our knowledge, no systematic study on \emph{event datasets} for SNNs.

\vspace{1mm}
\noindent\textbf{Motivation and Positioning of Our Work.}
Event streams encode information primarily in time, yet most dataset distillation methods are image-centric and rely on dense tensors or coarse binning, which mismatches sparse, polarity-signed, causal spikes and non-differentiable firing. This gap inflates storage and compute, complicates optimization, and weakens transfer to SNNs. We advocate an event-native, SNN-centric distillation paradigm that respects temporal structure with phase-aware alignment while remaining compatible with standard event-frame pipelines, enabling rapid SNN training under tight memory and latency.

\begin{figure*}[!t]
\centering
\includegraphics[width=\linewidth]{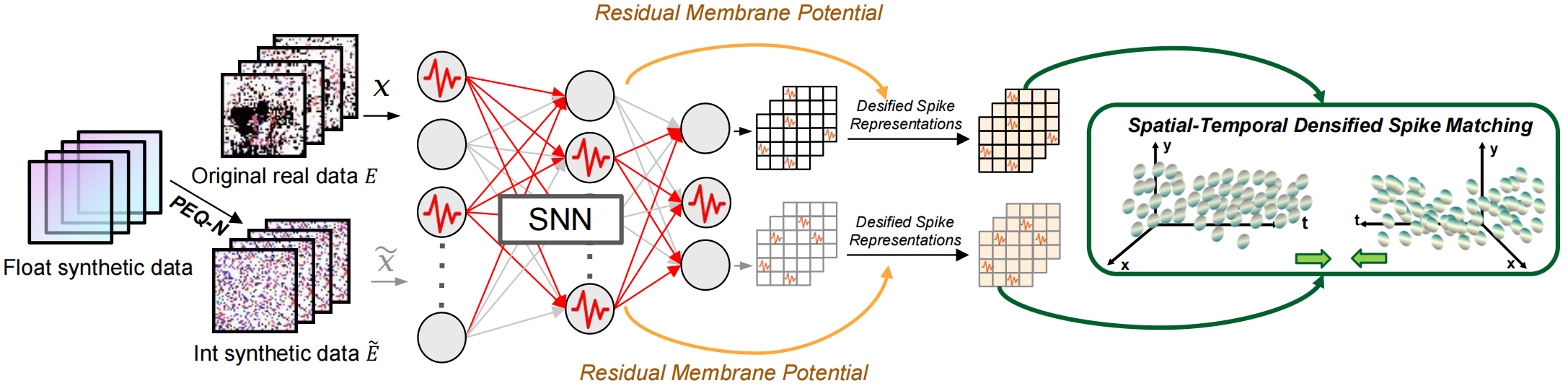}  
\vspace{-6mm}
\caption{The framework of our PACE (Phase-Aligned Condensation for Events). Float synthetic data is quantized into integer-driven events via our PEQ-N module. Then, both real and synthetic event streams are fed into unified SNN teacher and student models. The core of our approach is the Spatial-Temporal Densified Spike Matching module, which trains the synthetic data by forcing its resulting spike patterns to closely mimic those generated by the real data across both space and time.}
\label{fig3}
\end{figure*}

\section{Preliminary}
\label{sec:analysis-sd-lif}

\textbf{Neuron model.}
We briefly review the widely-used neuron model, \textit{i.e.,} the Leaky Integrate-and-Fire (LIF)~\citep{LIF}.
The membrane potential and spike firing of LIF model are given:
\begin{equation}\small
\begin{split}
H[t] \!&=\! V[t-1] \!+ \tau\!\left( X[t] \!-\! \left( V[t-1] \!-\! V_{\text{reset}} \right) \right),\\
S[t] \!&=\! \Theta\left( H[t] \!-\! V_{\text{th}} \right),~V[t] \!=\! H[t](1 \!-\! S[t]) \!+ V_{\text{reset}} S[t],
\end{split}
\label{eq:lif}
\end{equation}
where $\tau$ is the leaky factor ($\tau > 1$), $X[t]$ and $V[t]$ denote the input and membrane potential remaining at time step $t$. 
A binary spike $S[t]$ is emitted when the membrane potential $H[t]$ exceeds the threshold $V_{\text{th}}$. 
This is determined by the Heaviside function $\Theta(v)$, which outputs 1 if $v \geq 0$ and 0 otherwise.
After firing, the potential resets to $V_{\text{reset}}$. Otherwise, it remains at $H[t]$. 

\vspace{1mm}
\noindent\textbf{Event data.}
Event cameras (Dynamic Vision Sensors, DVS) offer microsecond-level temporal resolution, low latency, and a high dynamic range (\textgreater{}120\,dB), making them highly compatible with spiking neural networks (SNNs). Concretely, they asynchronously emit a positive or negative event at a pixel whenever the brightness change exceeds a threshold, forming a sparse event stream. Each event can be written as a 4-tuple
$\bm{e}_i = (t_i, x_i, y_i, p_i)$,
where $t_i$ is the timestamp, $(x_i,y_i)$ are the spatial coordinates, and $p_i$ is the polarity (1/0 or $1/-1$); $i$ indexes the $i$-th event in the stream. Although recent works explore alternative event representations~\citep{er1,er2,er3} to better exploit DVS characteristics, the strongest-performing pipelines to date still aggregate events into frames over fixed time windows. 
In our pipeline, we discretize time into $T$ bins and pack the raw stream into an event tensor $\bm{E}\in\mathbb{R}^{T\times C\times H\times W}$ (with $C\in\{1,2\}$ depending on whether polarities are merged or separated). With a finite bin width, this representation is inherently \emph{integer-driven} (per-bin counts); only in the theoretical asynchronous limit $T\!\to\!\infty$ (infinitesimal bins) does it converge to the native \emph{0--1 event-driven} process. To probe purely event-driven behavior and the theoretical case of distilling directly from streams, in our \texttt{bin} setting we disable within-bin accumulation and keep at most one spike per pixel per bin (0/1 occupancy).

\vspace{1mm}
\noindent\textbf{Dataset distillation via distribution matching.}
Because gradient-matching DC is slow and memory-hungry (and even more so for SNNs), we adopt distribution matching (DM) for efficiency. A classical DM objective aligns feature distributions produced by a fixed (or EMA-snapshotted) encoder $f(\cdot)$ under random augmentations $a\sim\mathcal{A}$:
\begin{equation}\small
\label{eq:dm}
\!\!\!\mathcal{L}_{\mathrm{DM}}(\tilde{\mathcal{X}})\!\!=\!\!
\mathop{\mathbb{E}}_{a\sim\mathcal{A}}
\!\Big\|\!
\underbrace{\tfrac{1}{|B|}\!\sum_{x\in B}\!\phi\!\big(f(a(x))\big)}_{\text{real features}}
\!-\!
\underbrace{\tfrac{1}{|\tilde{B}|}\!\sum_{\tilde{x}\in \tilde{B}}\!\phi\!\big(f(a(\tilde{x}))\big)}_{\text{synthetic features}}
\!\Big\|_2^2,
\end{equation}
where $B$ and $\tilde{B}$ are real and synthetic minibatches, and $\phi$ selects the feature type (\textit{e.g.,} activations, logits, or BN statistics)~\citep{zhao2022dm}.
Yet magnitude-only statistics can miss phase-sensitive structure. Neural Characteristic Function Matching (NCFM)~\citep{ncfm} matches the empirical characteristic functions (CFs) of features $z\!=\!f(x)$ over a frequency set $\Omega$:
\begin{equation}\small
\hat{\varphi}_z(\omega)\!=\!\tfrac{1}{B}\sum_{b=1}^{B} e^{\,\mathrm{i}\,\omega^\top z_b}, 
~\mathcal{L}_{\mathrm{CF}}\!=\!\sum_{\omega\in\Omega}\!\left\|
\hat{\varphi}_{z^{(\mathrm{syn})}}(\omega)\!-\!\hat{\varphi}_{z^{(\mathrm{real})}}(\omega)
\right\|_2^2,
\label{eq:ncfm}
\end{equation}
which constrains both magnitude and phase, capturing finer structure than magnitude-only matching.

%% file: sec/2_formatting.tex
\section{Method}
We present \emph{PACE} (Phase-Aligned Condensation for Events), a plug-and-play SNN dataset distillation framework that learns synthetic event data initialized from random noise. Sec.~\ref{sec:stdsm} introduces Spatial-Temporal Densified Spike Matching (ST-DSM), which densifies sparse spikes via residual-membrane-potential injection and aligns amplitude/phase using the Characteristic Function (CF) in the spatial domain and the Fast Fourier Transform (FFT) in the temporal domain. Sec.~\ref{sec:peq} presents \textbf{PEQ-N}, which outputs hard integer event frames in the forward pass and employs a straight-through estimator for gradients, while remaining compatible with standard event-frame pipelines.
The overall framework of PACE is in Fig.~\ref{fig3}. 


\subsection{Spatial-Temporal Densified Spike Matching}
\label{sec:stdsm}
\textbf{Spike features should be densified for phase-aware matching.}
In ANNs, intermediate features are continuous (\textit{e.g.,} ReLU activations), so CF-based matching can directly compare amplitudes and phases in feature space. In SNNs, however, layer-wise features are binarized spikes $s_t^{(l)}\!\in\!\{0,1\}^{D}$ and are extremely sparse in space \emph{and} time, which makes frequency-domain phase estimation brittle. To make this contrast explicit, let $a_t\!\in\!\mathbb{R}^{D}$ denote a continuous ANN feature and $s_t\!\in\!\{0,1\}^{D}$ a SNN feature. Their (per-time) characteristic functions (CFs) under a random projection $\omega\!\sim\!\mathcal{N}(0,I_D)$ are:
\begin{equation}\small
\Phi_a(\omega,t)\;=\;\mathbb{E}\!\left[e^{\,\mathrm{i}\,\omega^\top a_t}\right], 
\qquad
\Phi_s(\omega,t)\;=\;\mathbb{E}\!\left[e^{\,\mathrm{i}\,\omega^\top s_t}\right].
\label{eq:cf-contrast}
\end{equation}
Then we approximate $s_t$ by a Bernoulli vector with spike rate $\rho_t$ aligned with the all-ones direction:
\begin{equation}\small
\Phi_s(\omega,t)\approx\!(1-\rho_t)+\rho_t e^{\mathrm{i}\theta},
\quad\text{where}~ \theta=\omega^\top\mathbf{1}.
\label{eq:cf-binary}
\end{equation} When $\rho_t$ is small and spikes arrive irregularly, the phase $\arg\Phi_s(\omega,t)$ exhibits abrupt jumps tied to rare spike events, offering poor temporal-phase observability. By contrast, for continuous $a_t$, $\nabla_{a_t}\Phi_a(\omega,t)\!=\!\mathrm{i}\,\omega\,\Phi_a(\omega,t)$, so $\Phi_a(\omega,t)$ varies smoothly with $a_t$. Consequently, small drifts in $a_t$ induce a smoothly varying phase, which is the behavior exploited by CF-based phase matching.

\vspace{1mm}
\noindent\textbf{Densified spike representation (DSR) via residual membrane potential.}
To endow SNN features with a smooth temporal carrier while \emph{retaining explicit spike decisions}, we adopt the LIF notation in Eq.~\ref{eq:lif}: for layer $l$, the pre-reset membrane potential is $H^{(l)}[t]$, the spike is $S^{(l)}[t]=\Theta\!\big(H^{(l)}[t]-V^{(l)}_{\text{th}}\big)$, and $V^{(l)}_{\text{th}}$ is the threshold. We inject the \emph{residual (subthreshold) membrane potential} into the spike stream to obtain a \emph{single, continuous, spatio-temporally densified} feature per site:
\begin{equation}\small
\tilde S^{(l)}[t] \;=\; S^{(l)}[t] \;+\; \big(1 - S^{(l)}[t]\big)\,\frac{H^{(l)}[t]}{V^{(l)}_{\mathrm{th}}}
\;\in\; (-\infty,\,1].
\label{eq:densified}
\end{equation}
By Eq.~\ref{eq:lif}, when a spike occurs ($S^{(l)}[t]\!=\!1$), the potential resets. Otherwise ($S^{(l)}[t]\!=\!0$), it remains $H^{(l)}[t]$. Thus $\tilde S^{(l)}[t]\!=\!1$ at spike times and equals the normalized subthreshold value between spikes. Consequently, the CF of $\tilde S[t]$ admits a non-trivial subthreshold gradient even when no spike is fired:
\begin{equation}\small
\nabla_{H[t]}\,\mathbb{E}\!\left[e^{\,\mathrm{i}\,\omega^\top \tilde S[t]}\right]
\;=\;\frac{\mathrm{i}}{V_{\mathrm{th}}}
\mathbb{E}\!\left[
e^{\,\mathrm{i}\,\omega^\top \tilde S[t]}\;
\omega\odot\big(1-S[t]\big)
\right].
\label{eq:cf-dense-grad}
\end{equation}
We omit the layer superscript $(l)$. The mask $(1-S[t])$, from $S[t]=\Theta(H[t]-V_{\mathrm{th}})$ in Eq.~\ref{eq:lif}, restores phase sensitivity on subthreshold steps, which stabilizes temporal alignment for CF-based matching.

\vspace{1mm}
\noindent\textbf{CF in the spatial domain and FFT in the temporal domain for phase-aware matching.}
Event streams exhibit rich temporal dynamics, so purely spatial matching is insufficient. Leveraging the LIF dynamics in Eq.~\ref{eq:lif}, we propose \textbf{spatial-temporal densified spike matching (ST-DSM)}: combine CF-based projections in feature space with a discrete FFT along time to capture and align temporal phase. Formally, let $X,Y\in\mathbb{R}^{B\times T\times D}$ be densified features (Eq.~\ref{eq:densified}) from real and synthetic batches at a chosen layer. We sample $M$ random directions $\omega_m\sim\mathcal{N}(0,I_D)$ and compute empirical CFs at each time step ($t\in\{0,1,\dots,T-1\}, ~m\in\{0,1,\dots,M-1\}$):
\begin{equation}\small
\begin{aligned}
Z_r(m,t) &= \frac{1}{B}\sum_{b=0}^{B-1} \exp\!\big(\mathrm{i}\,\omega_m^\top X_{b,t,:}\big),\\
Z_s(m,t) &= \frac{1}{B}\sum_{b=0}^{B-1} \exp\!\big(\mathrm{i}\,\omega_m^\top Y_{b,t,:}\big).
\end{aligned}
\label{eq:cf}
\end{equation}

We then apply a full discrete Fourier transform along time (normalized with \texttt{forward}) to expose \emph{temporal phase}:
\begin{equation}\small
\begin{aligned}
F_r(m,\nu) &= \frac{1}{T}\sum_{t=0}^{T-1} Z_r(m,t)\,\mathrm{e}^{-\mathrm{i} 2\pi \nu t/T},\\
F_s(m,\nu) &= \frac{1}{T}\sum_{t=0}^{T-1} Z_s(m,t)\,\mathrm{e}^{-\mathrm{i} 2\pi \nu t/T}.
\end{aligned}
\label{eq:fft}
\end{equation}
where $\nu\in\mathcal{I}_T=\{0,1,\dots,T-1\}$ indexes temporal frequencies. Let $A_r=\lvert F_r\rvert$, $A_s=\lvert F_s\rvert$, and $\Delta\Phi(m,\nu)=\arg F_r(m,\nu)-\arg F_s(m,\nu)$. The ST-DSM loss at the chosen layer is
\begin{equation}\small
\!\!\begin{aligned}
\mathcal{L}^{(l)}_{\mathrm{ST\!-\!DSM}}
&= \frac{1}{TM}\sum_{m=0}^{M-1}\sum_{\nu\in\mathcal{I}_T}
\Big[
\alpha\big(A_r(m,\nu)\!-\!A_s(m,\nu)\big)^2 \\
&\!\!\!
+ 2\beta\,A_r(m,\nu)A_s(m,\nu)\big(1\!-\!\cos\Delta\Phi(m,\nu)\big)
\Big]^{\tfrac{1}{2}} .
\end{aligned}
\label{eq:stdsm}
\end{equation}
where $\alpha,\beta\!\in\![0,1]$ weight amplitude and phase terms. Because $\tilde S_t$ (Eq.~\ref{eq:densified}) mixes explicit spikes with a dense subthreshold trajectory, the temporal FFT (Eq.~\ref{eq:fft}) becomes sensitive to phase shifts that would be invisible on sparse $\{0,1\}$ spikes alone.

\vspace{1mm}
\noindent\textbf{Overall condensation loss.}
We combine ST-DSM with a time-expanded cross-entropy loss $\mathcal{L}_{\mathrm{CE}}$. Let $\{\mathbf{z}_t\!\in\!\mathbb{R}^{C}\}_{t=0}^{T\!-\!1}$ be the logits over time for a synthetic sequence. The condensation objective is
\vspace{-2mm}
\begin{equation}\small
\mathcal{L}_{\mathrm{condense}}
\;=\;
\lambda_{\mathrm{in}}\,
\mathcal{L}^{(l)}_{\mathrm{ST\!-\!DSM}}
\;+\;
\lambda_{\mathrm{inter}}
\mathcal{L}_{\mathrm{CE}}(\frac{1}{T}\sum_{t=0}^{T-1}\mathbf{z}_t,y),
\label{eq:total-overview}
\vspace{-2mm}
\end{equation}
where $l$ is set to be the last layer of feature extractor (before the linear layer). This objective is minimized only w.r.t.\ the synthetic data, and the teacher remains frozen during distillation.

\subsection{PEQ-$N$: Probabilistic Event Quantizer}
\label{sec:peq}
\textbf{Why DVS dataset distillation needs an event quantizer?}
Event streams are commonly converted into \emph{integer-valued event frames} by temporally binning raw binary events within short windows. This practice (i) aligns with standard training/evaluation pipelines and hardware interfaces that expect frame-like tensors; (ii) removes the arbitrary ordering of events inside a bin while preserving per-pixel counts; (iii) yields a bounded, discrete support $\{0,\dots,N\!-\!1\}$ that matches categorical modeling and stabilizes optimization; and (iv) keeps compatibility with inference-time representations used by off-the-shelf SNN backbones. To faithfully mirror this evaluation protocol, we impose the \textbf{same integer constraint} on our synthetic dataset.

\vspace{1mm}
\noindent\textbf{Discrete events with gradients.}
To keep integer event frames for inference \emph{and} retain gradients during distillation, we employ a probabilistic event quantizer, \emph{PEQ-$N$}. At each spatio-temporal location $r=(t,h,w,c)$, PEQ-$N$ predicts an $N$-way categorical distribution (temperature $\tau>0$) whose forward pass produces \emph{hard} integers,
while a straight-through estimator (STE) carries gradients so that the losses in Eq.~\ref{eq:total-overview} back-propagate to the data parameters.

\vspace{1mm}
\noindent\textbf{Formulation.}
Let $\mathbf{z}_r=(z_{r,0},\dots,z_{r,N-1})$ be logits for location $r$. We define
\begingroup\small
\begin{equation}\label{eq:peq-soft-hard}
\begin{aligned}
p_{r,n} &= \frac{\exp(z_{r,n}/\tau)}{\sum_{j=0}^{N-1}\exp(z_{r,j}/\tau)},\\
y^{\mathrm{soft}}_r &= \sum_{n=0}^{N-1} n\, p_{r,n}, \qquad
y^{\mathrm{hard}}_r &= \operatorname*{arg\,max}_{n} p_{r,n}.
\end{aligned}
\end{equation}
\endgroup
We combine discrete forward and continuous backward via straight-through estimator (STE):
\begin{equation}\small
y_r \;=\; y^{\mathrm{hard}}_r \;+\; \big(y^{\mathrm{soft}}_r - \mathrm{stopgrad}(y^{\mathrm{soft}}_r)\big).
\label{eq:ste}
\end{equation}
The gradients would flow through $y^{\mathrm{soft}}$. Using the softmax derivative, we have:
\begin{equation}\small
\frac{\partial p_{r,n}}{\partial z_{r,j}} \;=\; \frac{1}{\tau}\, p_{r,n} \big(\delta_{nj}-p_{r,j}\big),
\end{equation}
the analytic gradient of the soft expectation is given by:
\begingroup\small
\begin{equation}\label{eq:peq-grad}
\begin{aligned}
\frac{\partial y^{\mathrm{soft}}_r}{\partial z_{r,j}}
&= \sum_{n=0}^{N-1} n\, \frac{\partial p_{r,n}}{\partial z_{r,j}}\\
&= \frac{1}{\tau}\, p_{r,j}\!\left(j - \mathbb{E}_{p}[n]\right)
\;=\; \frac{1}{\tau}\, p_{r,j}\!\left(j - y^{\mathrm{soft}}_r\right).
\end{aligned}
\end{equation}
\endgroup
Because $\lvert j-\mathbb{E}_p[n]\rvert \le N-1$ and $0\le p_{r,j}\le 1$, the gradient is bounded:
\begin{equation}\small
\left\lvert \frac{\partial y^{\mathrm{soft}}_r}{\partial z_{r,j}} \right\rvert \;\le\; \frac{N-1}{\tau}.
\label{eq:peq-bound}
\end{equation}
For the binary special case $N=2$, $y^{\mathrm{soft}}_r=p_{r,1}$ and $\partial y^{\mathrm{soft}}_r / \partial z_{r,1}=(1/\tau)\,p_{r,1}(1-p_{r,1})$ with maximum $1/(4\tau)$. Compared with direct \texttt{round} or hard thresholding (almost everywhere zero gradient and undefined at jumps), Eq.~\ref{eq:peq-grad} and Eq.~\ref{eq:peq-bound} provide a smooth, bounded gradient channel; the temperature $\tau$ acts as an annealing knob for the smoothness-sharpness trade-off.

\vspace{1mm}
\noindent\textbf{Placement and coupling.}
PEQ-$N$ operates at the output of the synthetic data parameters:
$\hat{\mathbf{Y}} \xrightarrow{\mathrm{PEQ}\!-\!N} \hat{\mathbf{Y}}^{\mathrm{hard}}\!\in\!\{0,\dots,N\!-\!1\}^{T\times H\times W\times C}$.
The hard integers feed the frozen teacher to extract densified features $\tilde S$ (Eq.~\ref{eq:densified}) for the inner ST-DSM objective (Eq.~\ref{eq:stdsm}) and to produce time-varying logits for the discriminative term in the condensation loss (Eq.~\ref{eq:total-overview}). During backpropagation, gradients propagate through $y^{\mathrm{soft}}$ (Eq.~\ref{eq:ste}) into both the quantizer and the synthetic data parameters.

\section{Experiments}
\textbf{Implementation details.}
Experiments are conducted on NVIDIA A40 GPUs using the BrainCog platform~\citep{braincog}. For a fair comparison, We adopt the {VGGSNN}~\citep{deng2022tet} backbone across all experiments. 
For evaluation, we replicate existing coreset selection methods (\textit{Random}, \textit{Herding}\citep{herding}, \textit{K-Center}\citep{kcenter}) and dataset distillation baselines (\textit{DC}~\citep{Zhao2021GM}, \textit{DM}~\citep{zhao2022dm}, \textit{NCFM}~\citep{ncfm}) under spiking settings, to ensure a fair comparison with our method \textbf{PACE}.
PACE is designed to be plug-and-play and is integrated into the state-of-the-art NCFM pipeline. To improve efficiency, we significantly reduce the number of distillation iterations: for NCFM, we use only 5,000 iterations (1/4 of the original 20,000). All datasets are resized to an input resolution of $48\times48$. For CIFAR10-DVS and N-MNIST, we distill synthetic datasets with Images-Per-Class (IPC) values of 1, 10, and 50. For the smaller DVS-Gesture dataset, we adopt IPC values of 1, 5, and 10. The ratio of synthetic to full dataset size is reported in Table~\ref{tab:main}.
For the $M$ directions, we set it to 64 for all methods.

\vspace{1mm}
\noindent\textbf{Distillation strategies for different data types.}
We apply two tailored distillation strategies based on the input format:
\textbf{(1) Binary event data:} When the raw data consists of binary spikes, we also generate binary synthetic samples. We set the discretization parameter $N$ to 2 and use a high learning rate of 1.0 to encourage convergence to binary-like values (0 or 1).
\textbf{(2) Integer event data:} When the data contains integer spike counts, we set $N$ to 8 to allow richer quantization. A smaller learning rate $10^{-2}$ is used to facilitate finer-grained optimization of discrete-valued outputs.

\vspace{1mm}
\noindent\textbf{Evaluation protocol.}
We follow a rigorous evaluation setup to ensure stability and reproducibility. For each IPC setting, we run dataset distillation using 5 random seeds to produce 5 independent synthetic datasets. Each synthetic set is then used to train and evaluate 10 independently initialized models. We report the mean and standard deviation over the resulting 50 trials ($5 \times 10$), covering both training variance and synthetic data diversity.

\begin{table*}[t]
\centering
\caption{Dataset distillation results (Top-1 accuracy, \%). For all DD methods, synthetic sets are \emph{initialized from random noise} and optimized with a single VGGSNN backbone, which is also used to train and evaluate on the distilled data. Coreset Selection baselines operate on real samples. \textit{bin}/\textit{int} denote binary and integer event grids. IPC is images per class. Ratio is the fraction of synthetic data relative to the full training set. ``Full Dataset" trains the same VGGSNN on the entire real set. Results are reported as mean~$\pm$~std across runs.}
\vspace{-3mm}
\label{tab:main}
\setlength{\tabcolsep}{4pt}
\renewcommand{\arraystretch}{1}
\small
\begin{tabular}{cccccccccccc}
\toprule
\multirow{2}{*}{Dataset} & \multirow{2}{*}{Type} & \multirow{2}{*}{IPC} & \multirow{2}{*}{Ratio/\%}
& \multicolumn{3}{c}{Coreset Selection} & \multicolumn{4}{c}{Dataset Distillation (DD)}
& \multirow{2}{*}{Full Dataset} \\
\cmidrule(lr){5-7} \cmidrule(lr){8-11}
& & & & Random & Herding & K-Center & DC & DM & NCFM & PACE & \\
\midrule
\multirow{6}{*}{\rotatebox{90}{CIFAR10-DVS}}
  & \multirow{3}{*}{bin} & 1   & 0.1 & $16.2_{\pm0.2}$ & $18.9_{\pm0.3}$ & $10.3_{\pm0.3}$ & $19.9_{\pm0.6}$ & $15.0_{\pm0.9}$ & $23.1_{\pm0.1}$ & $\mathbf{27.3_{\pm1.2}}$ & \multirow{3}{*}{$53.4_{\pm0.5}$} \\
  &                       & 10  & 1   & $23.8_{\pm0.1}$ & $24.2_{\pm0.5}$ & $19.2_{\pm0.5}$ & $21.2_{\pm1.3}$ & $20.0_{\pm0.8}$ & $22.3_{\pm0.7}$ & $\mathbf{31.3_{\pm1.4}}$ & \\
  &                       & 50  & 5   & $34.7_{\pm0.4}$ & $32.0_{\pm0.3}$ & $30.1_{\pm0.3}$ & $19.8_{\pm1.6}$ & $20.6_{\pm1.0}$ & $24.6_{\pm0.6}$ & $\mathbf{41.9_{\pm0.5}}$  & \\
\cmidrule(lr){2-12}
  & \multirow{3}{*}{int} & 1   & 0.1 & $15.4_{\pm0.1}$ & $18.3_{\pm0.1}$ & $9.9_{\pm0.3}$  & $24.6_{\pm0.9}$ & $16.3_{\pm0.8}$ & $27.4_{\pm1.0}$ & $\mathbf{34.1_{\pm2.4}}$  & \multirow{3}{*}{$62.9_{\pm0.2}$} \\
  &                       & 10  & 1   & $24.1_{\pm0.2}$ & $25.4_{\pm0.3}$ & $19.9_{\pm0.5}$ & $32.2_{\pm1.3}$ & $25.0_{\pm0.9}$ & $28.9_{\pm3.1}$ & $\mathbf{41.3_{\pm1.9}}$  & \\
  &                       & 50  & 5   & $36.8_{\pm0.4}$ & $34.8_{\pm0.7}$ & $34.5_{\pm0.2}$ & $31.1_{\pm1.5}$ & $22.8_{\pm0.9}$ & $34.0_{\pm0.5}$ & $\mathbf{49.7_{\pm1.2}}$  & \\
\midrule
\multirow{6}{*}{\rotatebox{90}{N-MNIST}}
  & \multirow{3}{*}{bin} & 1   & 0.017 & $43.4_{\pm1.4}$ & $48.4_{\pm1.7}$ & $18.9_{\pm0.7}$ & $59.1_{\pm4.0}$ & $37.3_{\pm3.4}$ & $67.1_{\pm4.0}$ & $\mathbf{84.4_{\pm1.6}}$  & \multirow{3}{*}{$99.0_{\pm0.1}$} \\
  &                       & 10  & 0.17  & $84.7_{\pm0.2}$ & $79.8_{\pm0.6}$ & $82.6_{\pm1.0}$ & $62.4_{\pm15.2}$ & $46.5_{\pm5.0}$ & $87.2_{\pm4.0}$ & $\mathbf{91.8_{\pm0.8}}$  & \\
  &                       & 50  & 0.83  & $92.9_{\pm0.1}$ & $90.1_{\pm0.6}$ & $90.9_{\pm0.4}$ & $47.6_{\pm14.3}$ & $54.7_{\pm8.2}$ & $92.4_{\pm0.0}$ & $\mathbf{94.1_{\pm0.1}}$ & \\
\cmidrule(lr){2-12}
  & \multirow{3}{*}{int} & 1   & 0.017 & $58.4_{\pm0.6}$ & $63.6_{\pm0.2}$ & $18.7_{\pm0.4}$ & $63.1_{\pm5.5}$ & $47.8_{\pm6.4}$ & $71.8_{\pm2.1}$ & $\mathbf{84.7_{\pm0.5}}$  & \multirow{3}{*}{$99.3_{\pm0.0}$} \\
  &                       & 10  & 0.17  & $85.8_{\pm0.4}$ & $80.3_{\pm0.4}$ & $85.5_{\pm0.5}$ & $85.4_{\pm0.5}$ & $44.0_{\pm3.5}$ & $86.6_{\pm1.0}$ & $\mathbf{90.0_{\pm1.2}}$  & \\
  &                       & 50  & 0.83  & $93.9_{\pm0.1}$ & $89.8_{\pm0.3}$ & $91.3_{\pm0.2}$ & $91.1_{\pm1.7}$ & $46.5_{\pm7.5}$ & $89.8_{\pm0.6}$ & $\mathbf{94.8_{\pm0.1}}$  & \\
\midrule
\multirow{6}{*}{\rotatebox{90}{DVS-Gesture}}
  & \multirow{3}{*}{bin} & 1   & 0.93  & $28.0_{\pm1.9}$ & $40.0_{\pm0.6}$ & $14.1_{\pm1.1}$ & $40.4_{\pm4.9}$ & $21.8_{\pm2.3}$ & $33.6_{\pm1.3}$ & $\mathbf{50.7_{\pm2.1}}$ & \multirow{3}{*}{$75.5_{\pm0.8}$} \\
  &                       & 5   & 4.64  & $53.9_{\pm1.2}$ & $57.3_{\pm0.6}$ & $49.2_{\pm0.5}$ & $35.6_{\pm4.9}$ & $48.1_{\pm3.0}$ & $40.7_{\pm2.0}$ & $\mathbf{61.0_{\pm1.0}}$ & \\
  &                       & 10  & 9.29  & $58.8_{\pm1.6}$ & $58.8_{\pm1.5}$ & $60.4_{\pm1.1}$ & $35.9_{\pm5.7}$ & $48.3_{\pm3.3}$ & $48.7_{\pm2.9}$ & $\mathbf{68.7_{\pm1.7}}$ & \\
\cmidrule(lr){2-12}
  & \multirow{3}{*}{int} & 1   & 0.93  & $39.3_{\pm1.1}$ & $47.8_{\pm0.5}$ & $11.4_{\pm1.6}$ & $44.7_{\pm1.5}$ & $26.5_{\pm2.3}$ & $46.7_{\pm1.9}$ & $\mathbf{63.3_{\pm1.9}}$ & \multirow{3}{*}{$85.7_{\pm0.5}$} \\
  &                       & 5   & 4.64  & $52.8_{\pm0.8}$ & $56.2_{\pm1.3}$ & $51.2_{\pm0.5}$ & $53.7_{\pm1.7}$ & $26.6_{\pm2.4}$ & $48.0_{\pm2.1}$ & $\mathbf{70.3_{\pm2.5}}$ & \\
  &                       & 10  & 9.29  & $62.1_{\pm2.2}$ & $61.7_{\pm1.3}$ & $61.2_{\pm2.4}$ & $61.1_{\pm3.0}$ & $44.1_{\pm3.3}$ & $56.1_{\pm2.0}$ & $\mathbf{76.5_{\pm1.9}}$ & \\
\bottomrule
\end{tabular}%
\end{table*}

\subsection{Main Results}

\textbf{ANN-based DD methods fail to transfer.}
As shown in Table~\ref{tab:main}, distillation methods originally developed for RGB/frame-based ANNs, such as DC and the recent NCFM, do not maintain their advantage when directly applied to event-based {SNN-DVS} settings. For instance, on {DVS-Gesture} with \texttt{int} data at IPC=10, \textbf{NCFM} achieves only $56.1\%$, which is \emph{lower than all coreset methods}, including K-Center ($61.2\%$), Herding ($61.7\%$), and Random ($62.1\%$). Similar trends hold across other datasets and settings, showing that prior methods fail to model the spatio-temporal nature of event data.

\vspace{1mm}
\noindent\textbf{Our PACE lifts every baseline.}
Equipping prior DD with our \textbf{PACE} (NCFM+{PACE}) restores and \emph{amplifies} their advantage across \emph{all 18/18} settings in Table~\ref{tab:main}. Gains are especially pronounced on dynamic streams: on {DVS-Gesture} with \texttt{int} at {IPC=10}, NCFM climbs from $56.1\%$ to \textbf{76.5\%} (+20.4\%); with \texttt{bin} at {IPC=10}, it rises from $48.7\%$ to \textbf{68.7\%} (+20.0\%). Similar improvements appear on other datasets/budgets (\textit{e.g.,} {CIFAR10-DVS} \texttt{int}, {IPC=50}: $34.0\%\!\rightarrow\!\mathbf{49.7\%}$, +15.7\%; {N-MNIST} \texttt{bin}, {IPC=1}: $67.1\%\!\rightarrow\!\mathbf{84.4\%}$, +17.3\%), confirming the effectiveness of our methods.

\vspace{1mm}
\noindent\textbf{Coreset selection vs. \textbf{PACE}.}
Across datasets and budgets, learning \emph{synthetic} events outperforms selecting real ones. On {DVS-Gesture} with integer grids at the same IPC, the best coreset trails our distilled set by a large margin (\textit{e.g.,} at IPC=10, best coreset reaches only $61.2\%$, while ours achieves $76.5\%$). The gap stems from a fundamental capability difference: coresets directly select from the available real examples, while synthetic sets are \emph{learned} to reconstruct class prototypes and optimize for the downstream objective, which is crucial under tiny budgets.

\vspace{1mm}
\noindent\textbf{Where the gains concentrate.}
Benefits are largest for (i) \textbf{more dynamic} datasets ({DVS-Gesture} $\gg$ CIFAR10-DVS $\gtrsim$ N-MNIST) and (ii) \textbf{low-moderate budgets} (\(\mathrm{IPC}\!\le\!10\)). These regimes hinge on \emph{temporal shape} (onset/offset, rhythm, peak density). PACE densifies spikes via residual membrane potential and aligns amplitude/phase statistics (CF-in-feature, FFT-in-time), recovering precisely these shapes from very few sequences.


\vspace{1mm}
\noindent\textbf{Closeness to full-data performance.}
With only {IPC=10} on {DVS-Gesture} \texttt{int} \emph{(Ratio=\(9.29\%\) of the training set)}, the distilled set reaches \textbf{76.5\%}, about \(\sim\!89\%\) of the full-data upper bound (\(85.7\%\)).
A similar proximity holds on {N-MNIST} \texttt{int}, {IPC=10} \emph{(Ratio=\(0.17\%\))}: \(90.0\%\) vs.\ \(99.3\%\) (\(\sim\!91\%\)).
On {CIFAR10-DVS} \texttt{int}, {IPC=50} \emph{(Ratio=\(5\%\))}, performance reaches \(49.7\%\) vs.\ \(62.9\%\) (\(\sim\!79\%\)).


\vspace{1mm}
\noindent\textbf{Binary vs.\ integer event grids.}
At matched budgets, \texttt{int} \emph{tends} to outperform \texttt{bin}, especially on dynamic datasets ({DVS-Gesture}: {IPC=1} $63.3\%$ vs.\ $50.7\%$, {IPC=10} $76.5\%$ vs.\ $68.7\%$; {CIFAR10-DVS}: {IPC=10} $41.3\%$ vs.\ $31.3\%$). Integer multiplicity yields smoother pre-activation trajectories and stabler normalization after temporal/spatial aggregation, which reduces gradient variance and batch jitter during distillation/training. Binary grids are threshold-sensitive. Small misalignments get amplified by SNN nonlinearities. Our PACE partially mitigates this for \texttt{bin} by densifying spikes with residual membrane potential, yet \texttt{int} remains intrinsically advantageous on highly dynamic streams. A mild exception appears on {N-MNIST} at {IPC=10}, where \texttt{bin} slightly surpasses \texttt{int} (91.8\% vs.\ 90.0\%), likely due to simpler temporal structure where multiplicity adds limited benefit.

\subsection{How many time bins are needed?}

\begin{table}
\centering
\small
\setlength{\tabcolsep}{1pt}
\renewcommand{\arraystretch}{1.0}
\caption{Ablation of the time steps $T$ on \textbf{DVS-Gesture} (Accuracy \%) for our PACE. We choose $T\in\{2,4,6,8,10\}$. The ``Full data'' row reports bin/int performance as \textit{bin/int}.}
\vspace{-3mm}
\begin{tabular}{c|ccccc}
\toprule
Type & $T{=}2$ & $T{=}4$ & $T{=}6$ & $T{=}8$ & $T{=}10$\\
\midrule
Full data & $65.1/80.7$ & $75.5/85.7$ & $79.5/86.6$ & $80.3/87.1$ & $81.4/87.1$\\
\midrule
bin & $41.1_{\pm7.1}$ & $50.3_{\pm1.9}$ & $57.9_{\pm3.6}$ & $59.8_{\pm3.4}$ & $\mathbf{62.8_{\pm3.1}}$\\
int & $60.2_{\pm2.3}$ & $63.3_{\pm1.9}$ & $\mathbf{65.7_{\pm1.4}}$ & $64.6_{\pm1.9}$ & $65.5_{\pm1.1}$\\
\bottomrule
\end{tabular}
\label{tab:step-ablation}
\end{table}
\textbf{More $T$ helps until saturation.}
On \textbf{DVS-Gesture} (Table~\ref{tab:step-ablation}), \texttt{bin} improves near-monotonically with $T$, from $41.1\%$ at $T{=}2$ to $62.8\%$ at $T{=}10$ (+21.7 \%). In contrast, \texttt{int} peaks at $T{=}6$ ($65.7\%$) and changes only marginally thereafter ($64.6\%$ at $T{=}8$, $65.5\%$ at $T{=}10$). The full-data upper bound similarly saturates around $T{\ge}8$ for \texttt{int} (about $87\%$), indicating diminishing returns once dominant temporal rhythms are captured.

\vspace{1mm}
\noindent\textbf{Why \texttt{int} saturates earlier.}
Integer encodes timing and intensity per bin, which smooths pre-activations and stabilizes normalization under temporal and spatial aggregation. This lowers gradient variance and batch jitter, so moderate $T$ is sufficient. Binary grids lack amplitude cues and are threshold-sensitive, thus they benefit more from higher temporal resolution. PACE densifies spikes and partly compensates for \texttt{bin}, yet \texttt{int} remains intrinsically easier to optimize at moderate $T$.

\subsection{Ablation study of PEQ-N and ST-DSM in PACE}

\begin{table}
\centering
\small
\setlength{\tabcolsep}{4pt}
\renewcommand{\arraystretch}{1.0}
\caption{Ablation of codebook size $N$ in \textbf{PEQ-N} on \textbf{DVS-Gesture} (Accuracy \%) for our PACE. }
\vspace{-3mm}
\begin{tabular}{c|ccccc}
\toprule
Type & $N{=}2$ & $N{=}4$ & $N{=}8$ & $N{=}16$ & $N{=}32$\\
\midrule
bin & $\mathbf{50.7_{\pm2.1}}$ & $49.0_{\pm3.7}$ & $49.1_{\pm2.9}$ & $47.5_{\pm4.9}$ & $46.9_{\pm6.5}$\\
int & $51.7_{\pm2.5}$ & $54.9_{\pm1.7}$ & $\mathbf{63.3_{\pm1.9}}$ & $59.3_{\pm2.1}$ & $59.9_{\pm1.9}$\\
\bottomrule
\end{tabular}
\label{tab:knum-ablation}
\end{table}
\textbf{Effect of $N$ in PEQ-N.} We conduct experiments on DVS-Gesture as shown in Table~\ref{tab:knum-ablation}.
For \texttt{int}, accuracy improves from $51.7\%$ ($N{=}2$) to $54.9\%$ ($N{=}4$) and peaks at $63.3\%$ with $N{=}8$, indicating that a moderate codebook preserves multiplicity without over-fragmenting counts. Larger $N$ brings no further gain and tends to create sparse bins that weaken gradients and stability. For \texttt{bin}, performance declines as $N$ grows ($50.7\!\rightarrow\!49.0\!\rightarrow\!49.1\!\rightarrow\!47.5\!\rightarrow\!46.9$), consistent with sharper quantization boundaries interacting poorly with thresholded inputs and amplifying jitter. In practice, use $N{\approx}8$ for \texttt{int} and keep $N$ small ($2$) for \texttt{bin}.

\begin{table}
\centering
\small
\setlength{\tabcolsep}{4pt}
\renewcommand{\arraystretch}{1.0}
\caption{Ablation of ST-DSM on \textbf{DVS-Gesture} (Accuracy \%). }
\vspace{-3mm}
\begin{tabular}{c||ccc | ccc}
\toprule
\multirow{2}{*}{IPC}&\multicolumn{3}{c|}{\textbf{integer grid}} & \multicolumn{3}{c}{\textbf{binary grid}} \\
\cmidrule(lr){2-4}\cmidrule(lr){5-7}
& DSR & ST-SM & Acc  & DSR & ST-SM & Acc \\
\midrule
\multirow{4}{*}{1}  & \ding{55} & \ding{55} & $46.7_{\pm1.9}$   & \ding{55} & \ding{55} & $40.8_{\pm5.1}$ \\
  & \ding{51} & \ding{55} & $56.5_{\pm2.7}$   & \ding{51} & \ding{55} & $47.6_{\pm7.0}$ \\
  & \ding{55} & \ding{51} & $46.9_{\pm7.4}$   & \ding{55} & \ding{51} & $47.7_{\pm5.0}$ \\
  & \ding{51} & \ding{51} & $\mathbf{63.3_{\pm1.9}}$  & \ding{51} & \ding{51} & $\mathbf{50.3_{\pm2.1}}$ \\
\midrule
\multirow{4}{*}{5}  & \ding{55} & \ding{55} & $48.0_{\pm2.1}$  & \ding{55} & \ding{55} & $48.1_{\pm2.1}$ \\
  & \ding{51} & \ding{55} & $60.3_{\pm3.1}$ & \ding{51} & \ding{55} & $55.1_{\pm2.9}$ \\
  & \ding{55} & \ding{51} & $50.8_{\pm2.3}$ & \ding{55} & \ding{51} & $52.1_{\pm2.0}$ \\
  & \ding{51} & \ding{51} & $\mathbf{70.3_{\pm2.5}}$ &  \ding{51} & \ding{51} & $\mathbf{61.0_{\pm1.0}}$ \\
\midrule
\multirow{4}{*}{10} & \ding{55} & \ding{55} & $52.7_{\pm1.7}$ & \ding{55} & \ding{55} & $54.7_{\pm1.7}$ \\
 & \ding{51} & \ding{55} & $63.7_{\pm1.1}$ & \ding{51} & \ding{55} & $55.1_{\pm2.9}$ \\
 & \ding{55} & \ding{51} & $55.3_{\pm2.8}$ & \ding{55} & \ding{51} & $54.6_{\pm4.4}$ \\
 & \ding{51} & \ding{51} & $\mathbf{76.5_{\pm1.9}}$ &  \ding{51} & \ding{51} & $\mathbf{68.7_{\pm1.0}}$ \\
\bottomrule
\end{tabular}
\label{tab:module-ablation-side}
\end{table}

\vspace{1mm}
\noindent\textbf{Ablation of ST-DSM.} We conduct ablation of the ST-DSM's components on DVS-Gesture as shown in Table~\ref{tab:module-ablation-side}. 
{DSR} denotes the densified spike representation in Eq.~(9). Disabling ST-SM removes the FFT alignment, i.e., Eq.~(10) $\rightarrow$ Eq.~(4). 
We can conclude: \textbf{(i):} {DSR} is the primary driver and stabilizer: for \texttt{int} it lifts accuracy from \(46.7/48.0/52.7\) to \(56.5/60.3/63.7\) (IPC \(1/5/10\)), whereas \textbf{(ii):} ST-SM alone is smaller or volatile (\(46.9/50.8/55.3\)). \textbf{(iii):} Combining {DSR+ST-SM} (ST-DSM) is consistently best, \texttt{int}: \(63.3/70.3/76.5\); \texttt{bin}: \(50.3/61.0/68.7\), and also reduces variance (\textit{e.g.,} \texttt{bin} IPC\(=1\), std \(2.1\) vs \(5\!\sim\!7\)).

\begin{figure*}[t]
\centering
\includegraphics[width=\linewidth]{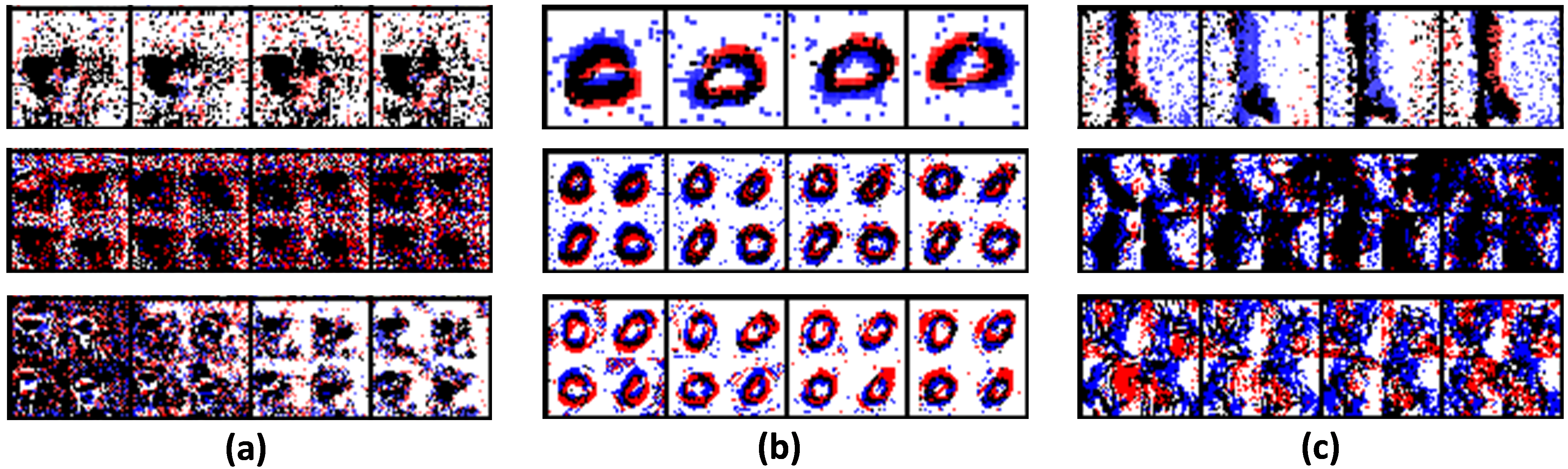}  
\vspace{-5mm}
\caption{Visualization of Original real data (top), distilled binary (middle) and integer (bottom) event data for (a) DVS-Gesture ``right hand wave", (b) N-MNIST ``0", and (c) CIFAR10-DVS ``airplane". Each subfigure shows representative voxelized event maps: red denotes positive (ON) events; blue denotes negative (OFF) events; black marks pixels where both polarities occur within the same bin.}
\label{fig1}
\vspace{-2mm}
\end{figure*}

\subsection{Computational and Storage Efficiency}
\label{sec:energy_cost}

Table~\ref{tab:summary_dd_full} summarizes the efficiency of different dataset distillation methods in terms of performance, training speed (s/iter), storage, and GPU memory. 
All efficiency metrics (time, storage, and memory) are measured on a single NVIDIA A40 (48 GB), so time and memory serve as direct proxies for computational and energy cost, while the number of stored images reflects disk footprint. More detailed analysis can be found in sec \ref{efficiency} of Supplementary material.

\newcommand{\oom}{\textcolor{red}{OOM}}  
\newcommand{\best}[1]{\textbf{#1}}


\begin{table}[t]
\centering
\small
\setlength{\tabcolsep}{3pt}
\renewcommand{\arraystretch}{1.0}
\caption{\textbf{Comparison of DD methods on N-MNIST (IPC=1).}
We report test accuracy (\%), training speed (second per iteration), disk storage of the train set (number of images), and peak GPU memory usage (GB). 
All efficiency metrics (time, storage, and memory) are measured on a single NVIDIA A40 (48 GB), and the whole dataset serves as a non-condensed reference.}
\vspace{-3mm}
\begin{tabular}{lcccc|c}
\toprule
& \textbf{DC} & \textbf{DM} & \textbf{NCFM} & \textbf{PACE} & \textbf{Whole Data} \\
\midrule
Performance & [59.1] & [37.3] & [67.1] & \best{[84.4]} & [99.0] \\
Training Speed   & [8.7] & \best{[2.5]} & [3.7] & [3.7] & [192.5] \\
Storage  & [10] & [10] & [10] & [10] & [60000] \\
GPU Memory& [4.5] & [4.4] & [2.8] & \best{[2.8]} & [2.8] \\
\bottomrule
\end{tabular}
\label{tab:summary_dd_full}
\end{table}

\subsection{Discussion}
\textbf{Visualization analysis.}
Fig.~\ref{fig1} compares real events (top) with distilled \texttt{bin} (middle) and \texttt{int} (bottom) for
(a) DVS\mbox{-}Gesture ``right hand wave'',
(b) N\mbox{-}MNIST ``0'',
and (c) CIFAR10\mbox{-}DVS ``airplane''.
Across classes, \texttt{int} shows clearer contours and more coherent temporal evolution than \texttt{bin}:
alternating ON/OFF bands track the hand wave, a smooth annulus forms for the digit ``0'',
and elongated motion edges appear for the airplane instead of salt\mbox{-}and\mbox{-}pepper noise.
Mixed\mbox{-}polarity pixels cluster on true edges for \texttt{int}, while \texttt{bin} is sparser and more flicker\mbox{-}prone.
Both distilled variants suppress background clutter relative to real data, and \texttt{int} better preserves fine structure, consistent with its higher accuracy.
Overall, PACE yields compact, class\mbox{-}consistent spatiotemporal templates, with \texttt{int} offering stronger shape and rhythm fidelity.
More detailed visualization and analysis can be found in Sec \ref{vis} in Supplementary materials.

\begin{table}
\centering
\small
\setlength{\tabcolsep}{3pt}
\renewcommand{\arraystretch}{1.0}
\caption{Across-architecture generalization on \textbf{DVS-Gesture}. Synthetic data are distilled with VGGSNN and then used to train/evaluate other backbones.}
\vspace{-3mm}
\begin{tabular}{c c | c c c}
\toprule
Method & Grid & VGGSNN & SNN-ConvNet & SEW-ResNet \\
\midrule
\multirow{2}{*}{NCFM} & bin & $33.6_{\pm2.1}$ & $46.8_{\pm3.6}$ & $12.4_{\pm2.3}$ \\
 & int & $46.7_{\pm1.9}$ & $54.8_{\pm2.7}$ & $30.5_{\pm1.9}$ \\
 \midrule
\multirow{2}{*}{PACE} & bin & $50.7_{\pm2.1}$ & $62.5_{\pm0.6}$ & $29.6_{\pm0.6}$ \\
 & int & $63.3_{\pm1.9}$ & $68.2_{\pm1.4}$ & $37.7_{\pm1.9}$ \\
\bottomrule
\end{tabular}
\label{tab:transfer}
\vspace{-2mm}
\end{table}

\vspace{1mm}
\noindent\textbf{Transferability.}
We explore whether data distilled on VGGSNN can transfer well to distinct SNN backbones, as shown in Table~\ref{tab:transfer}. Compared to NCFM, our method improves SNN-ConvNet by \textbf{+17.1} \% (\texttt{bin}: $62.5$ vs.\ $46.8$) and \textbf{+13.4} \% (\texttt{int}: $68.2$ vs.\ $54.8$); SEW-ResNet by \textbf{+17.2} \% (\texttt{bin}: $29.6$ vs.\ $12.4$) and \textbf{+7.2} \% (\texttt{int}: $37.7$ vs.\ $30.5$); and the source VGGSNN by \textbf{+17.1} (\texttt{bin}) and \textbf{+16.6} (\texttt{int}) points. The consistent gains across \texttt{bin}/\texttt{int} suggest PACE learns spatiotemporal statistics not tied to a specific architecture. Accuracy on SEW-ResNet is lower, hinting at a larger inductive gap for residual/gated dynamics. Extending transfer to residual families (including spiking ResNets) with residual-aware alignment is promising future work.


\section{Conclusion}
In this paper, we introduce \textbf{PACE}, the first dataset distillation framework designed for fast SNN training on event streams. 
PACE combines two core components: \textbf{ST-DSM} and \textbf{PEQ-N}.
ST-DSM uses residual membrane potentials to densify spike-based features (SDR) and performs fine-grained spatiotemporal matching of amplitude and phase (ST-SM),
while PEQ-N is a plug-and-play straight through probabilistic integer quantizer.
Across DVS-Gesture, CIFAR10-DVS, and N-MNIST, PACE consistently outperforms existing coreset selection and dataset distillation baselines, with the largest gains on dynamic event streams and at low or moderate IPC. 
In particular, on N-MNIST with IPC\(=1\), PACE achieves \(84.4\%\) accuracy, which is about \(85\%\) of the \(99.0\%\) obtained with the full training set, while reducing training time by over \(50\times\) and storage cost by \(6000\times\).
The distilled sets transfer to other SNN backbones, and the overall event native pipeline reduces storage and wall clock training time, enabling minute scale convergence on neuromorphic streams and supporting efficient edge deployment.
